\documentclass{article}
\pdfoutput=1

\usepackage[preprint]{neurips_2022}

\usepackage[utf8]{inputenc} 
\usepackage[T1]{fontenc}    
\usepackage{hyperref}       
\usepackage{url}            
\usepackage{booktabs}       
\usepackage{amsfonts}       
\usepackage{nicefrac}       
\usepackage{microtype}      
\usepackage{xcolor}         
\usepackage{amsmath}
\usepackage{pdfpages}

\title{MAFormer: A Transformer Network with Multi-scale Attention Fusion for Visual Recognition}

\author{
    Yunhao Wang$^{1,\dagger}$, Huixin Sun$^{2,\dagger}$ , Xiaodi Wang$^{1}$, Bin Zhang$^{1}$ , Chao Li$^{1}$, \\
    \textbf{Ying Xin$^{1}$, Baochang Zhang$^{2,\ast }$, Errui Ding$^{1}$, Shumin Han$^{1,\ast}$}\\
    $^{1}$Department of Computer Vision Technology (VIS), Baidu Inc\\
    $^{2}$Beihang University, Beijing, China\\
    $^{\ast}$Corresponding author, email: \texttt{bczhang@buaa.edu.cn}, \texttt{hanshumin@baidu.com}\\
    $^{\dagger}$Equal contributions\\
}

\begin{document}

\maketitle

\begin{abstract}
 Vision Transformer and its variants have demonstrated great potential in various computer vision tasks. But conventional vision transformers often focus on global dependency at a coarse level, which suffer from a learning challenge on global relationships and fine-grained representation at a token level. In this paper, we introduce Multi-scale Attention Fusion into transformer (\textbf{MAFormer}), which explores local aggregation and global feature extraction in a dual-stream framework for visual recognition. We develop a simple but effective module to explore the full potential of transformers for visual representation by learning fine-grained and coarse-grained features at a token level and dynamically fusing them. Our Multi-scale Attention Fusion (MAF) block consists of: i) a local window attention branch that learns short-range interactions within windows, aggregating fine-grained local features; ii) global feature extraction through a novel Global Learning with Down-sampling (GLD) operation to efficiently capture long-range context information within the whole image; iii) a fusion module that self-explores the integration of both features via attention. Our MAFormer achieves state-of-the-art performance on common vision tasks. In particular, MAFormer-L achieves 85.9$\%$ Top-1 accuracy on ImageNet, surpassing CSWin-B and LV-ViT-L by 1.7$\%$ and 0.6$\%$ respectively. On MSCOCO, MAFormer outperforms the prior art CSWin by 1.7\% mAPs on object detection and 1.4\% on instance segmentation with similar-sized parameters, demonstrating the potential to be a general backbone network.
\end{abstract}

\section{Introduction}
Transformers have prevailed in computation vision since the breakthrough of ViT~\cite{dosovitskiy2020image}, attaining excellent results in various visual tasks, including image recognition, object detection, and semantic segmentation. Despite these progress, the global self-attention mechanism in line with ViT~\cite{dosovitskiy2020image,li2021improved} has a quadratic computation complexity to the input image size, which is insufferable for high-resolution scenes. To reduce the complexity, several variants have been introduced to replace global self-concern with local self-concern. Swin Transformer~\cite{liu2021swin} with a hierarchical architecture partitions input features into non-overlapping windows and shifts the window positions by layer. After that various window partition mechanisms are designed for better local feature capturing. CSWin Transformer~\cite{dong2021CSWin} splits features into horizontal and vertical stripes in parallel, aiming to enlarge the window receptive field. However, it only focuses on the information within windows, leaving the dependencies across windows unexplored. Shuffle Transformer~\cite{huang2021shuffle} revisits the  ShuffleNet~\cite{ma2018shufflenet} and embeds the spatial shuffle in local windows to intensify their connections. While these local window-based attention methods have achieved excellent performance, even better than the convolutional neural network  (CNN) counterparts (e.g., ResNets~\cite{he2016deep,szegedy2017inception}), they suffer from a learning challenge on the global relationship that is indispensable for a  better feature representation.

Another line of research efforts focuses on combining CNNs with transformers, which are trade-offs between local patterns and global patterns. CvT~\cite{zhang2020feature} transforms the linear projection in the self-attention block into convolution projection. CoatNet~\cite{yan2021contnet} merges depth-wise convolution with self-attention via simple relative attention and stacks convolution and attention layers in a principled way. DS-Net~\cite{mao2021dual} proposes a dual-stream framework that fuses convolution and self-attention via cross-attention, where each form of scale learns to align with the other. However, as shown in DS-Net~\cite{mao2021dual}, convolution and attention hold intrinsically conflicting properties that might cause ambiguity in training. For instance, the long-range information captured by global self-attention could perturb the neighboring details of convolution in high-resolution feature maps, compromising both global and local representations.

In this paper, we develop a Multi-scale Attention Fusion transformer (\textbf{MAFormer}), which explores local aggregation and global feature extraction in a dual-stream transformer framework. To avoid the risk of incompatibility between convolution and self-attention, we apply local window attention to extract fine-grained feature representation. We also design a Global Learning with Down-sampling (GLD) module to extract global features, which captures coarse-grained features based on the full-sized input. We further encode token-level location information of the input into global representations via positional embeddings. Moreover, we describe two dual-stream architectures based on different fusion strategies, particularly the Multi-scale Attention Fusion (MAF) scheme that can fully explore the potential of both features. Its effectiveness can be explained by the fact that MAF block can enhance the interaction between each local-global token pair, where local features and global features are co-trained in a unified framework, formulating a more ample and informative representation. The contributions of this work are concluded as follows.
\begin{enumerate}

\item A MAFormer network is introduced to extract and fuse fine-grained and coarse-grained features at a token level, which can self-explore  the integration of both features via attention  to improve the representation capacity for the input image.
\item A local window attention branch is first introduced to learn the short-range interactions within local windows.  We further introduce a Global Learning with Down-sampling (GLD) module on the dual branch, which efficiently captures the long-range context information within the whole image.
\item We develop two dual-stream architectures based on different fusion strategies, particularly the Multi-scale Attention Fusion (MAF) scheme that can fully explore the potential of both features.
\item Without bells and whistles, the proposed MAFormer outperforms prior vision Transformers by large margins in terms of recognition performance. We also achieve state-of-the-art results over the previous best CSWin for object detection and instance segmentation with similar parameters.
\end{enumerate}

\begin{figure}
  \centering
  \includegraphics[scale=0.45,page=1]{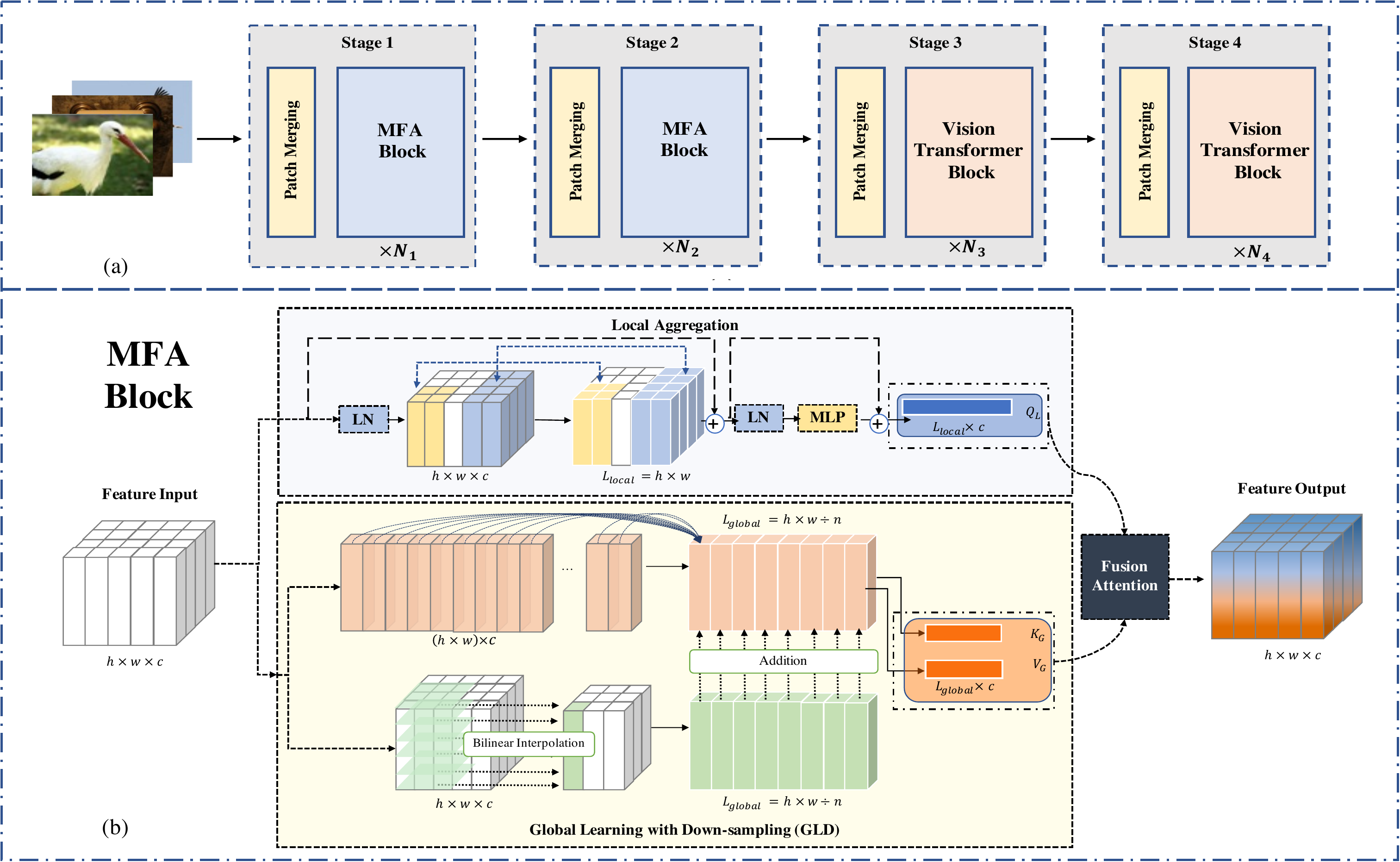}
  \caption{Architecture of MAFormer. We utilize the MAF block in the first two stages, which incorporates a Local Aggregation branch and a Global Learning with Down Sampling (GLD) branch. Both streams are fed into a fusion module to improve the capability of feature representation. }
  \label{fig_framework}
\end{figure}

\section{Related Work}
\paragraph{Vision transformers.} Self-attention-based architectures, in particular Transformers~\cite{vaswani2017attention}, have become the dominant model for Natural Language Processing (NLP). Motivated by its success in NLP, ViT~\cite{dosovitskiy2020image} innovatively applies a pure-transformer architecture to images by splitting an image into patches and equating them with tokens (words). ViT shows the competitive effect on classification tasks~\cite{deng2009imagenet}. Many efforts have been devoted to applying ViT for various vision tasks, including object detection~\cite{carion2020end,zhu2020deformable,yao2021efficient,wang2021pnp,roh2021sparse}, semantic segmentation~\cite{cheng2021per, strudel2021segmenter, xie2021segformer} , pose estimation~\cite{li2021tokenpose, yang2020transpose, yuan2021hrformer}, re-identification~\cite{he2021transreid}, and low-level image processing~\cite{chen2021pre}. These results verify the outstanding ability of the transformer as a general visual backbone. However, the self-attention mechanism is inefficient to encode low-level features, hindering their high potential for efficient representation learning.

\paragraph{Local window attention-based transformers.} Vision transformers demonstrate a high capability in modeling the long-range dependencies, which is especially helpful for handling high-resolution inputs in downstream tasks. However, such methods adopt the original full self-attention and their computational complexity is quadratic to the image size. To reduce the cost, some recent vision Transformers~\cite{liu2021swin, vaswani2021scaling} adopt the local window self-attention mechanism~\cite{ramachandran2019stand} and its shifted/haloed version that adds the interaction across different windows. To enlarge the receptive field, axial self-attention~\cite{ho2019axial} and criss-cross attention~\cite{huang2019ccnet} propose calculating attention within stripes along horizontal or/and vertical axis instead of fixing local windows as squares. Inspired by axial self-attention and criss-cross attention, the method~\cite{dong2021CSWin} presents the Cross-Shaped Window self-attention. CSWin performs the self-attention calculation in the horizontal and vertical stripes in parallel, with each stripe obtained by splitting the input feature into stripes of equal width.

\paragraph{Convolution in transformers.} According 
to recent analysis~\cite{peng2021conformer, dai2021coatnet}, convolution networks and transformers hold different merits. While the convolution operation guarantees a better generalization and fast convergence, thanks to its inductive bias, attention formulate networks with higher model capacity. Therefore, combining convolutional and attention layers can joint these advantages and achieve better generalization and capacity at the same time. Some existing transformers explore the hybrid architecture to incorporate both operations for better visual representation. Comformer~\cite{peng2021conformer} proposes the Feature Coupling Unit to fuse convolutional local features with transformer-based global representations in an interactive fashion. CvT~\cite{wu2021cvt} designs convolutional token embedding and convolutional transformer block for capturing more precise local spatial context. Followingly, CoatNet~\cite{dai2021coatnet} merges depth-wise convolution into attention layers with simple relative attention. Apart from incorporating explicit convolution, some works~\cite{liu2021swin, dong2021CSWin, yuan2021tokens, wang2021pyramid} try to incorporate some desirable properties of convolution into the Transformer backbone.

\section{Method}
\subsection{Overall Architecture}\label{para:framework}
The Multi-scale Attention Fusion  mechanism is proposed to extract fine-grained and coarse-grained features at a token level and fuse them dynamically, which formulates a general vision transformer backbone, dubbed as MAFormer, improving the performance in various visual tasks. Fig.~\ref{fig_framework}(a) shows the overall architecture of MAFormer. It takes an image $\mathcal{X} \in {R}^{H \times W \times 3}$ as input, where ${W}$ and ${H}$ represents the width and height of the input image, and employs a hierarchical design. By decreasing the resolution of feature maps, the network captures multi-scale features across different stages. We partition an input image into patches and perform patch merging, receiving $\frac{H}{4} \times \frac{W}{4}$ visual tokens with $C$ feature channels. From there, the tokens flow through two stages of MAF Blocks and the two stages of the original Vision Transformer Blocks. Within each stage, MAFormer adopts a patch merging layer by convention which downsamples the spatial size of the feature map by 2$\times$, while the feature channel dimension is increased. 

According to recent studies into feature representations~\cite{raghu2021vision}, visual transformers like the ViT attend locally and globally in its lower layers but primarily focus on global information in higher layers. In light of the pattern, we incorporate multi-scale feature representations in the first two stages of MAFormer, while in the last two stages, the original vision transformer block is utilized, where the resolution of the features is reduced and the computational cost of full attention becomes affordable.

\subsection{Multi-scale Attention Fusion Block}\label{para:attn}

\label{para:MAFormer}
\begin{figure}[!t]
  \centering
  \includegraphics[scale=0.3,page=1]{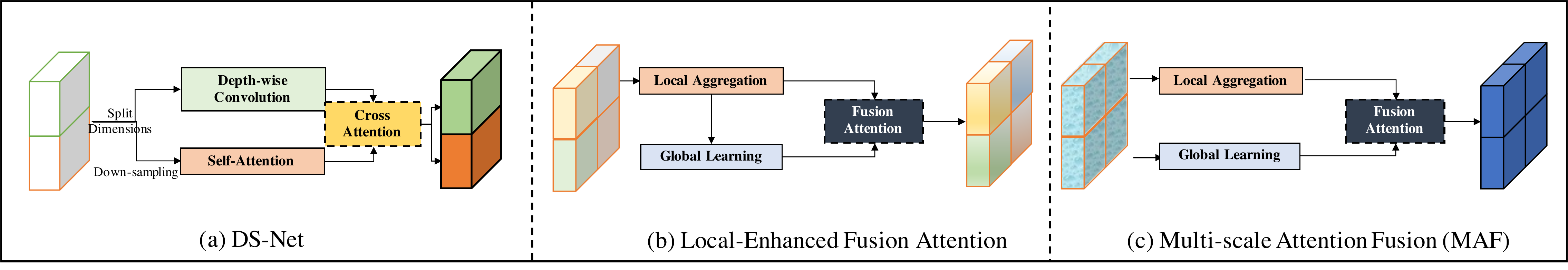}
  \caption{Different designs in dual-stream multi-scale representations.}
  \label{block}
\end{figure}

In this section, we elaborate the details of our Multi-scale Attention Fusion (MAF) block. As shown in Fig.~\ref{fig_framework}{(b)}, the MAF block includes a Local Aggregation branch and a Global Learning with Down Sampling (GLD) branch, generating token-level fine-grained and coarse-grained features respectively. Both streams are fed into a fusion module to improve the capability of feature representation. 

\paragraph{Local aggregation.}Previous hybrid networks~\cite{dai2021coatnet,li2022uniformer} utilize CNNs to extract local features, which are further integrated into a Transformer branch. Yet, such approaches risk the mismatch between convolution and self-attention. In MAF, we avoid the incompatibility and explore the usage of local window-based multi-head attention mechanisms as the fine-grained representation. Considering an input $X\in \textbf{R}^{H \times W \times C}$, the local aggregation $X_{L}^{l}$ is defined as: 

\begin{align}
&{X_{L}^{l}} = \text{Local-Window-Attention}\left( {\text{LN}\left({X^{l-1}} \right)} \right) + {X}^{l-1} ,\nonumber\\
&{X_{L}^{l}} = \text{MLP}\left({X_{L}^{l}} \right) + {X_{L}^{l}},
\end{align}

where ${{X}}^l$ denotes the output of $l$-th Transformer block.

\paragraph{Global feature extraction.}
Although local window self-attention methods have achieved excellent performance, they can only capture window-wise information and fail to explore the dependencies across them. Also, existing methods are still challenged in global dependency extraction due to insufficient usage of coarse-grained contextual information. As such, efficient capture of the global dependencies is constitutive for model representation. 

To address these issues, we introduce a Global Learning with Down-sampling (GLD) module to extract global information from a large-sized input. To this end, we first utilize a single neuron layer that is fully connected to the feature input. Without cutting out any dimensions, it output a down-sampled contextual abstraction that is dynamically learned. As illustrated in the Fig.~\ref{fig_framework}(c), the input $X\in \textbf{R}^{H \times W \times C}$ is first flattened to $X_G\in \textbf{R}^{C \times L}$, where $L$ is equal to $H \times W$. Then $X_G\in \textbf{R}^{C \times L}$ is globally extracted by a fully connected layer, downsized to scaling ratio $N$. During experiments, we have tuned several values of $N$ and 0.5 is optimal, which is set as the default in MAFormer. Further, we encode the token-level location information of the input into global representations via positional embeddings. As illustrated in the Fig. ~\ref{fig_framework}(c), the $Pos$ operation utilizes a layer-wise bilinear interpolation as the measure and $FC$ represents as the full connection.
\begin{align}
    &{X_{G}^{l}} = Pos\left(X_{G}^{l-1}\right) + FC(X_{G}^{l-1}),
\end{align}

where $X_{G}^{l}$ denotes the global branch output of $l$-th Transformer block.

\paragraph{Multi-scale attention fusion (MAF).} We develop two types of dual-stream multi-scale representations, as shown in Fig.~\ref{block}. First, we extract global dependencies on top of local representations as an enhancement, aiming to provide information flow across local windows. As shown in Fig.~\ref{block}(b), the GLD module takes the output of local window attention and fuses the global representations back with local. However, such approach can only capture the global correlations between local attributes, not from input. Therefore, we propose the Multi-scale attention fusion (MAF) measure, extracting the local and global scales of input directly and  separately. Both stream of information are fed into a fusion block via attention, as shown in Fig.~\ref{block}(c). In this way, our MAF block can capture the correlations between each local-global token pair and prompts the local features to adaptively explore their relationship with global representation, enabling themselves to be more representative and informative. 


Given extracted local features $X_{L} \in \textbf{R}^{C \times L_{local}}$ and global features $X_{G} \in \textbf{R}^{C \times L_{global}}$, the Multi-scale Attention Fusion is defined as:

\begin{equation}
	\label{co-attention-qkv}
	\begin{aligned} 
	 &Q_L = X_{L}W_{Q}^{local},  \\
	 &K_G = X_{G}W_{K}^{global},  \\
     &V_G = X_{G}W_{V}^{global},
	\end{aligned}
\end{equation}
where $W_{Q}^{local}$, $W_{K}^{global}$, $W_{V}^{global}$ are learning hyper-parameter matrix. Then we calculate the Multi-scale Attention Fusion  (MAF) between every pair of $X_{L}$ and $X_{G}$:
\begin{equation}
	\label{co-attention-weight-1}
	\begin{aligned}
	\text{MAF}\Big(Q_L,K_G,V_G \Big) = \operatorname{softmax}(\frac{Q_L K_G^T}{\sqrt{d}})V_G.
	\end{aligned}
\end{equation}

\section{Experiment}
\begin{table}
  \caption{Detailed settings of MAFormer of different model sizes and their performance on ImageNet-1k validation set. In all configurations, the expansion ratio of each MLP is set as 4.}
  \label{table1}
  \centering
    \begin{tabular}{ccccccc}
    \hline
    Models & Dim & Blocks & Params(M) &FLOPs(G) &Top1(\%)  \\\hline 
    \hline
    MAFormer-S & [64, 128, 320, 512] & [3, 5, 8, 3] & 23 & 4.5 & 83.7\\
    MAFormer-B & [64, 128, 320, 512] & [3, 8, 20, 7] & 53 & 9.8 & 85.0\\
    MAFormer-L & [128, 192, 448, 640] & [3, 8, 24, 7] & 104 & 22.6 & 85.9\\
    \hline
    \end{tabular}
\end{table}

In this section, we first provide ablation studies of the MAF block. Then, we give the experimental results of MAFormer in three settings: image classification, object detection with instance segmentation and semantic segmentation. Specifically, we use ImageNet-1K~\cite{deng2009imagenet} for classification, MSCOCO 2017~\cite{lin2014microsoft} with Mask R-CNN~\cite{he2017mask} and Cascade R-CNN~\cite{cai2018cascade} for object detection with instance segmentation, and ADE20K~\cite{zhou2017scene} for semantic segmentation, where we employ the semantic FPN~\cite{2019Panoptic} and UPerNet~\cite{xiao2018unified} as the basic framework. All experiments are conducted on V$100$ GPUs.

\subsection{Ablation Study and Analysis}
The multi-scale attention fusion (MAF) module in MAFormer network is mainly a composition of three: the Local Aggregation mechanism, the Global Learning with Down-sampling (GLD) module, and the dynamic fusion module. In the following experiments, we explore the best-performed structure of MAFormer by substituting and ablating different parts of the network. We set MAFormer-S as the baseline and all experiments are conducted on the image classification dataset ImageNet-1K.

\paragraph{Local aggregation.} The selection of attention method in the Local Aggregation module is very flexible, which could be substituted by different approaches on window based self-attention~\cite{huang2019ccnet,liu2021swin,ho2019axial}. In the MAF block, we compare the original work on window partition~\cite{liu2021swin} and its recent variant cross-shaped window based self-attention~\cite{dong2021CSWin}. As shown as Table~\ref{table-ablation-LG}, the experiments demonstrate that MAFormer-S using the cross-shaped window based self-attention outperforms shifted window-based self-attention by +0.2$\%$ top-1 accuracy on ImageNet $1$K, which is set as the default approach.

\paragraph{Global feature extraction.}\label{para:global_exp}  Global information is vital to feature representation. We show in Table~\ref{table-ablation-LG} that MAFormer-S with GLD yields +1\% top-1 accuracy than methods without global information on ImageNet-$1$K. We also compare GLD with other measures that extract global information and down-sample the input at the same time. As shown, GLD brings +0.3\% accuracy than basic-configured convolution, demonstrating that the detailed information from global tokens can be extracted in a learnable and dynamic manner using GLD, with local positional information encoded.

\paragraph{Fusion structure analysis.} 
The implementations of different connection modules are compared in Table~\ref{table-ablation-fusion}. As shown, our proposed Multi-scale Fusion Attention is more efficient than the previous local/global dual-stream architecture~\cite{mao2021dual}. Also, MAF is validated in our experiments with +0.2\% superiority over the local enhanced fusion measure. Instead of fixing the fusion, cross-scale information transfers are automatically determined by feature themselves, making the combined more effective.

\setlength{\tabcolsep}{2pt}
\begin{table}
\begin{center}
  \caption{Ablation study of different local aggregation and global feature representation modules.}
  \label{table-ablation-LG}
  \centering
    \begin{tabular}{cccccc}
    \hline
    Method & Params & Attention in Local Aggregation & Global Feature Extraction &Top1 (\%)\\ \hline \hline
    Swin-T & 29M & Shifted Window~\cite{liu2021swin} & None & 81.3 \\ \hline
    CSWin-T  & 23M & Cross-shaped~\cite{dong2021CSWin} & None & 82.7 \\ \hline
    MAFormer-S & 23M & Shifted Window~\cite{liu2021swin} & GLD & 83.4 \\ \hline    
    MAFormer-S & 23M & Cross-shaped~\cite{dong2021CSWin} &  Convolution & 83.4 \\ \hline
    \textbf{MAFormer-S} & \textbf{23M} &  \textbf{Cross-shaped~\cite{dong2021CSWin}}  & \textbf{GLD} &\textbf{83.7} \\ \hline
    \end{tabular}
\end{center}
\end{table}

\setlength{\tabcolsep}{1pt}
\begin{table}
    \caption{Accuracy of MAFormer-S using different structure design.}
    \label{table-ablation-fusion}
    \centering
        \begin{tabular}{cccc}
        \hline
        Framework & Params(M)  & Dual Stream Design & Top1 (\%)\\ 
        \hline \hline
        DS-Net~\cite{mao2021dual} & 23M & Co-Attention from Convolution and Self-attention & 82.3 \\ \hline
        MAFormer-S & 23M & Local-Enhanced Fusion Attention & 83.5 \\ \hline
        \textbf{MAFormer-S} & \textbf{23M} & \textbf{Multi-scale Fusion Attention} &\textbf{83.7} \\ \hline
        \end{tabular}
\end{table}

\subsection{Image Classification on ImageNet-1K}
\paragraph{Settings.} 
In this section, we conduct experiments of MAFormer on ImageNet-$1$K classification~\cite{deng2009imagenet} and compare the proposed architecture with the previous state-of-the-arts. MAFormer follows~\cite{jiang2021all} by default and is trained with Token Labeling~\cite{jiang2021all}. Dropout regularization rate~\cite{srivastava2014dropout} is set as $0.1$/$0.3$/$0.4$ for MAFormer-S/B/L respectively, as shown in Table~\ref{table1}. The learning rate of MAFormer-S and MAFormer-B are $1.6$e-$3$, while for MAFormer-L it is $1.2$e-$3$. All experiments are conducted on V$100$ GPUs.

\paragraph{Results.} As shown in Table~\ref{table1}, MAFormer-S with only $23$M parameters can achieve a top-$1$ accuracy of $83.7$\% on ImageNet-$1$k. Increasing the embedding dimension and network depth can further boost the performance. Table~\ref{table2} shows in details that MAFormer outperforms the previous state-of-the-art vision transformers. Specifically, MAFormer-L achieves 85.9$\%$ Top-1 accuracy with 22.6G FLOPs, surpassing CSWin-B~\cite{dong2021CSWin} and LV-ViT-L~\cite{jiang2021all} by 1.7$\%$ and 0.6$\%$ respectively. MAFormer variants also outperform the prior art hybrid architectures~\cite{dai2021coatnet,mao2021dual} and local window-attention-based transformers~\cite{huang2021shuffle,2021CrossViT,liu2021swin} by large margins with a fair amount of computation.

\begin{table}
  \caption{Comparison with the state-of-the-art on ImageNet-1K. $\dagger$ indicates with Token Labeling~\cite{jiang2021all}.}
  \label{table2}
  \centering
    \begin{tabular}{l|cccc|c}
    \toprule
    \hline
    Models & Train Size & Test Size & Params(M) & FLOPs(G) & Top1($\%$) \\ \hline
    \midrule
    DeiT-S ~\cite{2020Training} & $224^2$ & $224^2$ & 22 & 4.6 & 79.8 \\
    Swin-T ~\cite{liu2021swin} & $224^2$ & $224^2$ & 29 & 4.5 & 81.3 \\
    CrossViT-15 ~\cite{2021CrossViT} & $224^2$ & $224^2$ & 27 & 5.8 & 81.5 \\
    CoAtNet-0 ~\cite{dai2021coatnet}& $224^2$ & $224^2$ & 25 & 4.6 & 81.6 \\
    Focal-T ~\cite{2021Focal} & $224^2$ &$224^2$ & 29 & 4.9 & 82.2 \\
    DS-Net-S ~\cite{mao2021dual} & $224^2$ & $224^2$ & 23 & 3.5 & 82.3 \\
    Shuffle-T ~\cite{huang2021shuffle} & $224^2$ & $224^2$ & 29 & 4.6 & 82.5 \\
    CSWin-T ~\cite{dong2021CSWin} & $224^2$ & $224^2$ & 23 & 4.3 & 82.7 \\
    \textbf{MAFormer-S}  & $224^2$ & $224^2$ & 23 & 4.5 & \textbf{83.0} \\
    LV-ViT-S$\dagger$ ~\cite{jiang2021all} & $224^2$ & $224^2$ & 26 & 6.6 & 83.3 \\
    \textbf{MAFormer-S$\dagger$} & $224^2$ & $224^2$ & 23  & 4.5 & \textbf{83.7}\\ \hline
    \midrule
    CrossViT-18 ~\cite{2021CrossViT} & $224^2$ & $224^2$ & 44 & 9.5 & 82.8 \\
    Swin-S ~\cite{liu2021swin} & $224^2$ & $224^2$ & 50 & 8.7 & 83.0 \\
    DS-Net-B ~\cite{mao2021dual} & $224^2$ & $224^2$ & 49 & 8.4 & 83.1 \\
    Twins-SVT-B ~\cite{chu2021twins} & $224^2$ & $224^2$ & 56 & 8.3 & 83.2 \\
    CoAtNet-1 ~\cite{dai2021coatnet} & $224^2$ & $224^2$ & 42 & 8.4 & 83.3 \\
    Shuffle-S ~\cite{huang2021shuffle} & $224^2$ & $224^2$ & 50 & 8.9 & 83.5 \\
    Focal-S ~\cite{2021Focal} & $224^2$ & $224^2$ & 51 & 9.1 & 83.5 \\
    CSWin-S ~\cite{dong2021CSWin} & $224^2$ & $224^2$ & 35  & 8.9 & 83.6\\
    LV-ViT-M$\dagger$ ~\cite{jiang2021all} & $224^2$ & $224^2$ & 56 & 16 & 84.1 \\
    \textbf{MAFormer-B$\dagger$} &  $224^2$ & $224^2$ & 53 & 9.8 & \textbf{85.0}\\\hline
    \midrule
    DeiT-B ~\cite{2020Training} & $224^2$ & $224^2$ & 86 & 17.5 & 81.8 \\
    CrossViT-B ~\cite{2021CrossViT} & $224^2$ & $224^2$ & 105 & 21.2 & 82.2 \\
    Swin-B ~\cite{liu2021swin} & $224^2$ & $224^2$ & 88 & 15.4 & 83.5 \\
    Focal-B ~\cite{2021Focal} & $224^2$ & $224^2$ & 90 & 16.0 & 83.8 \\
    Shuffle-B ~\cite{huang2021shuffle} & $224^2$ & $224^2$ & 88 & 15.6 & 84 \\
    CSWin-B ~\cite{dong2021CSWin} & $224^2$ & $224^2$ & 78 & 15.0 & 84.2 \\
    CoAtNet-3 ~\cite{dai2021coatnet} & $224^2$ & $224^2$ & 168 & 34.7 & 84.5 \\
    CaiT-M36 ~\cite{2021Going} & $224^2$ & $384^2$ &271 & 247.8 & 85.1 \\
    LV-ViT-L$\dagger$ ~\cite{jiang2021all} & $288^2$ & $288^2$ & 150 & 59.0 & 85.3 \\
    \textbf{MAFormer-L$\dagger$} & $224^2$ & $224^2$ & 105 & 22.6 & \textbf{85.9}\\\hline
    \bottomrule
    \end{tabular}
\end{table}

\subsection{Object Detection and Instance Segmentation on MSCOCO}

\setlength{\tabcolsep}{2.5pt}
    \begin{table}[ht]
    \begin{center}
        \caption{Object detection and instance segmentation performance on the COCO val2017 with the Mask R-CNN framework. The FLOPs (G) are measured at resolution 800x1280, and the models are pretrained on the ImageNet-1K.}
        \label{table5}
        \resizebox{100mm}{!}{
        \begin{tabular}{l|c|c|c|c|c|c|c|c}
            \toprule
            \hline
            Backbone & Params  & FLOPs   &  \multicolumn{6}{c}{Mask R-CNN 1x schedule} \\ 
            ~ & (M) & (G) & $AP^b$ & $AP^b_{50}$ & $AP^b_{75}$ & $AP^m$ & $AP^m_{50}$ & $AP^m_{75}$ \\ \hline
            \midrule
            Res50 ~\cite{he2016deep} & 44 & 260 & 38.0 & 58.6 & 41.4 & 34.4 & 55.1 & 36.7\\ 
            PVT-S ~\cite{wang2021pyramid} & 44 & 245 & 40.4 & 62.9 & 43.8 & 37.8 & 60.1 & 40.3 \\
            ViL-S ~\cite{zhang2021multi} & 45 & 218 & 44.9 & 67.1 & 49.3 & 41. & 64.2 & 44.1\\
            TwinsP-S ~\cite{chu2021twins} & 44 & 245 & 42.9 & 65.8 & 47.1 & 40.4 & 62.7 & 42.9 \\
            Twins-S ~\cite{chu2021twins} & 44 & 228 & 43.4 & 66.0 & 47.3 & 40.3 & 63.2 & 43.4  \\
            Swin-T ~\cite{liu2021swin} & 48 & 264 & 42.2 & 64.6 & 46.2 & 39.1 & 64.6 & 42.0 \\
            CSWin-T ~\cite{dong2021CSWin} & 42 & 279 & 46.7 & 68.6 & 51.3 & 42.2 & 65.6 & 45.4 \\
            \textbf{MAFormer-S} & 41 & 256 & \textbf{47.0} & \textbf{69.5} & \textbf{51.6} & \textbf{42.7} & \textbf{66.5} & \textbf{46.1} \\ \hline
             \midrule
            Res101 ~\cite{he2016deep} & 63 & 336 & 40.4 & 61.1 & 44.2 & 36.4 & 57.7 & 38.8  \\
            X101-32 ~\cite{xie2017aggregated} & 63 & 340 & 41.9 & 62.5 & 45.9 & 37.5 & 59.4 & 40.2 \\
            PVT-M ~\cite{wang2021pyramid} & 64 & 302 & 42.0 & 64.4 & 45.6 & 39.0 & 61.6 & 42.1 \\
            ViL-M ~\cite{zhang2021multi} & 60 & 261 & 43.4 & -- & -- & 39.7 & -- & --   \\
            TwinsP-B ~\cite{chu2021twins} & 64 & 302 & 44.6 & 66.7 & 48.9 & 40.9 & 63.8 & 44.2  \\
            Twins-B ~\cite{chu2021twins} & 76 & 340 & 45.2 & 67.6 & 49.3 & 41.5 & 64.5 & 44.8 \\
            Swin-S ~\cite{liu2021swin} & 69 & 354 & 44.8 & 66.6 & 48.9 & 40.9 & 63.4 & 44.2 \\
            CSWin-S ~\cite{dong2021CSWin} & 54 & 342 & 47.9 & 70.1 & 52.6 & 43.2 & 67.1 & 46.2\\
            \textbf{MAFormer-B} & 71 & 354 & \textbf{49.6} & \textbf{71.4} & \textbf{54.7} & \textbf{44.6} & \textbf{68.6} & \textbf{48.4} \\\hline
            \midrule
            X101-64 ~\cite{xie2017aggregated} & 101 & 493 & 42.8 & 63.8 & 47.3 & 38.4 & 60.6 & 41.3  \\
            PVT-L ~\cite{wang2021pyramid} & 81 & 364 & 42.9 & 65.0 & 46.6 & 39.5 & 61.9 & 42.5  \\
            ViL-B ~\cite{zhang2021multi} & 76 & 365 & 45.1 & -- & -- & 41.0 & -- & --  \\
            TwinsP-L ~\cite{chu2021twins} & 81 & 364 & 45.4 & -- & -- & 41.5 & -- & --  \\
            Twins-L ~\cite{chu2021twins} & 111 & 474 & 45.9 & -- & -- & 41.6 & -- & --  \\ 
            Swin-B ~\cite{liu2021swin} & 107 & 496 & 46.9 & -- & -- & 42.3 & -- & -- \\
            CSWin-B ~\cite{dong2021CSWin} & 97 & 526 & 48.7 & 70.4 & 53.9 & 43.9 & 67.8 & 47.3\\
            \textbf{MAFormer-L} & 122 & 609 & \textbf{50.7} & \textbf{72.4} & \textbf{55.6} & \textbf{45.4} & \textbf{69.7} & \textbf{49.2}  \\ \hline
            \bottomrule 
            \end{tabular}
            }
    \end{center}
\end{table}

\begin{table}[ht]
\begin{center}
\setlength{\tabcolsep}{3pt}
\caption{Object detection and instance segmentation performance on the COCO val2017 with the Cascade R-CNN framework. The FLOPs (G) are measured at resolution 800x1280, and the models are pretrained on the ImageNet-1K.}
\label{table:det1}
\vspace{0.3cm}

\resizebox{100mm}{!}{
\begin{tabular}{l|c|c|c|c|c|c|c|c}
    \hline
    \toprule
    Backbone & Params  &FLOPs   &  \multicolumn{6} {c}{Cascade R-CNN 3x schedule}\\
    ~ & (M) & (G) & $AP^b$ & $AP^b_{50}$ & $AP^b_{75}$ & $AP^m$ & $AP^m_{50}$ & $AP^m_{75}$ \\
    \midrule 
    Res50 ~\cite{he2016deep} & 82 & 739 & 46.3 & 64.3 & 50.5 & 40.1 & 61.7 & 43.4 \\
    Swin-T ~\cite{liu2021swin} & 86 & 745 & 50.5 & 69.3 & 54.9 & 43.7 & 66.6 & 47.1 \\
    CSWin-T ~\cite{dong2021CSWin} & 80 & 757 & 52.5 & \textbf{71.5} & 57.1 & 45.3 & 68.8 & 48.9 \\
    \textbf{MAFormer-S} & 80 & 733 & \textbf{52.6} & 71.3 & \textbf{57.3} & \textbf{45.7} & \textbf{68.9} & \textbf{49.8} \\
    \midrule
    X101-32 ~\cite{xie2017aggregated} & 101 & 819 & 48.1 & 66.5 & 52.4 & 41.6 & 63.9 & 45.2 \\
    Swin-S ~\cite{liu2021swin} & 107 & 838 & 51.8 & 70.4 & 56.3 & 44.7 & 67.9 & 48.5 \\
    CSWin-S ~\cite{dong2021CSWin} & 92 & 820 & 53.7 & 72.2 & 58.4 & 46.4 & 69.6 & 50.6 \\
    \textbf{MAFormer-B} & 109 & 833 & \textbf{54.4} & \textbf{72.8} & \textbf{59.2} & \textbf{46.8} & \textbf{70.4} & \textbf{51.0} \\
    \midrule
    X101-64 ~\cite{xie2017aggregated} & 140 & 972 & 48.3 & 66.4 & 52.3 & 41.7 & 64.0 & 45.1 \\
    Swin-B ~\cite{liu2021swin} & 145 & 982 & 51.9 & 70.9 & 56.5 & 45.0 & 68.4 & 48.7 \\
    CSWin-B ~\cite{dong2021CSWin} & 135 & 1005 & 53.9 & 72.6 & 58.5 & 46.4 & 70.0 & 50.4 \\
    \textbf{MAFormer-L }& 160 & 1088 & \textbf{54.7} & \textbf{73.2} & \textbf{59.4} & \textbf{47.3} & \textbf{71.2} & \textbf{51.3} \\
    \hline
    \bottomrule 
\end{tabular}
}
\vspace{-0.5cm}
\end{center}
\end{table}

Pre-training models on image classification resources and adapting them to downstream tasks has become the standard approach in most vision works. However, the data volume of downstream tasks is much lower than classification benchmarks, the ImageNet for instance. According to recent studies~\cite{raghu2021vision}, the lower layers of attention-based networks perform poorly on aggregating local correlations when trained a small amount of data, given the lack of inductive bias. As a result, state-of-the-art transformer backbones on the ImageNet provide no significant improvement to downstream subtasks. MAFormer, on the other hand, utilize local window based attention in the lower layers and strategically encode global information with it. In this way, local patterns are easier to acquire when the training data is not sufficient, making it a general and efficient visual backbone.

\paragraph{Settings.} To demonstrate the merits of MAFormer on downstream tasks, we evaluate the model on COCO object detection task~\cite{lin2014microsoft}. We first utilize the typical framework Mask R-CNN~\cite{he2017mask}, where we configure 1x schedule with 12 epochs training schedules. In details, the shorter side of the image is resized to 800 while keeping the longer side no more than 1333. We utilize the same AdamW~\cite{loshchilov2017decoupled} optimizer with initial learning rate of 1e-4, decayed by 0.1 at epoch 8 and 11(1x schedule), and weight decay of 0.05. We set stochastic drop path regularization of 0.2 for MAFormer-S backbone, and 0.3 for MAFormer-B and MAFormer-L backbone, referred in Table~\ref{table1}.

To extend our research, we evaluate MAFormer in another typical framework Cascade R-CNN~\cite{cai2018cascade}. For Cascade R-CNN, we adopt 3x schedule with 36 epochs training schedules and the multi-scale training strategy~\cite{carion2020end, sun2021sparse} to randomly resize the shorter side between 480 to 800. We utilize the same AdamW~\cite{loshchilov2017decoupled} optimizer with initial learning rate of 1e-4, decayed by 0.1 at epoch 27 and 33, and weight decay of 0.05. We set stochastic drop path regularization of 0.2, 0.3, and 0.4 for MAFormer-S, MAFormer-B, and MAFormer-L backbone, respectively.

We compare MAFormer with various works: typical CNN backbones ResNet~\cite{he2016deep}, ResNeXt~\cite{xie2017aggregated}, and competitive Transformer backbones PVT~\cite{wang2021pyramid}, Twins~\cite{chu2021twins},  Swin~\cite{liu2021swin} and CSWin~\cite{dong2021CSWin}.

\paragraph{Results.} Table~\ref{table5} reports box mAP ($AP^b$) and mask mAP ($AP^s$) of the Mask R-CNN framework with 1x training schedule. It shows that the MAFormer variants notably outperform all the CNN and Transformer counterparts. Our MAFormer-S, MAFormer-B, and MAFormer-L achieve 47.0\%, 49.6\%, and 50.7\% box mAP for object detection, surpassing the previous best CSWin Transformer by +0.3\%, +1.7\%, and +2.0\%. Besides, our models present consistent improvement in instance segmentation, with +0.5\%, +1.4\%, and +1.5\% mask mAP higher than the previous best backbone. Notably, MAFormer-B outperforms CSWin-S and Swin-S with far less parameters.

Table~\ref{table:det1} contains the box mAP ($AP^b$) and mask mAP ($AP^m$) results from the Cascade R-CNN framework with 3x training schedule. It shows that MAFormer variants outperform all the CNN and Transformer counterparts in great margin. Specifically, MAFormer-S, MAFormer-B, and MAFormer-L achieve 52.6\%, 54.4\%, and 54.7\% box mAP for object detection, surpassing the previous best CSWin Transformer by +0.1\%, +0.7\%, and +0.8\%. Besides, our variants also have consistent improvement on instance segmentation, which are +0.3\%, +0.4\%, and +0.9\% mask mAP higher than the previous best backbone. It shows with a stronger framework, MAFormer still surpass the counterparts by promising margins under different configurations.

\subsection{Experiments of Semantic Segmentation with semantic FPN and UPerNet on ADE20K}

\paragraph{Settings.}ADE20K~\cite{zhou2017scene} is a widely used semantic segmentation dataset, covering a broad range of 150 semantic categories. It has 25K images in total, with 20K for training, 2K for validation, and another 3K for testing. We further investigate the capability of MAFormer for semantic segmentation on the ADE20K dataset. Here we employ the semantic FPN~\cite{2019Panoptic}and UPerNet~\cite{xiao2018unified} as the basic framework. All experiments are conducted on 8 V100 GPUs. For fair comparison, we train Semantic FPN~\cite{2019Panoptic} 80k iterations with batch size as 16, and UPerNet~\cite{xiao2018unified} 160k iterations with the batch size as 16 and the image  resolution is 512×512.

\paragraph{Results.}In Table~\ref{table:seg}, we provide the experimental results in terms of mIoU and Multi-scale tested mIoU (MS mIoU). 

It shows that MAFormer-S, MAFormer-B achieve 47.9, 49.8 with the semantic FPN framework, 6.4 and 2.6 higher mIoU than the Swin-Transformer~\cite{liu2021swin}. Also, MAFormer-S, MAFormer-B achieve 49.8, 51.1 with the UPerNet framework, 3.9, 3.0 higher mIoU than the Swin-Transformer~\cite{liu2021swin}. 

\begin{table}
\begin{center}
\caption{Comparison with previous best results on ADE20K semantic segmentation. UPerNet: learning rate of 6 × $10{^{-5}}$, a weight decay of 0.01, a scheduler that uses linear learning rate decay, and a linear warmup of 1,500 iterations. Semantic FPN: learning rate of 2 × $10{^{-4}}$, a weight decay of 1 × $10{^{-4}}$, a scheduler that uses Cosine Annealing learning rate decay, and a linear warmup of 1,000 iterations. The FLOPs are measured at resolution 2048×512.}
\label{table:seg}
\vspace{0.3cm}
\resizebox{120mm}{!}{
\begin{tabular}{l|c|c|c|c|c|c|c}
\hline

\toprule
Models &  \multicolumn{3} {c|}{Semantic FPN 80K} & \multicolumn{4} {c}{UPerNet 160k}\\
 ~  &$\#$Params(M) & FLOPs(G) & mIoU($\%$)  & $\#$Params(M) & FLOPs(G) & mIoU($\%$) & MS mIoU($\%$) \\
\midrule
Res50~\cite{he2016deep} & 29 & 183 & 36.7 & - & - & - & - \\
Twins-S ~\cite{2021Twins} & 28 & 144 & 43.2 & 54 & 901 & 46.2 & 47.1\\
TwinsP-S ~\cite{2021Twins} & 28 & 162 & 44.3 & 55 & 919 & 46.2 & 47.5\\
Swin-T ~\cite{liu2021swin} & 32 & 182 & 41.5 & 60 & 945 & 44.5 & 45.8\\
Focal-T ~\cite{2021Focal} & - & - & -  & 62 & 998 & 45.8 & 47.0 \\
Shuffle-T ~\cite{huang2021shuffle} & - & - & -  & 60 & 949 & 46.6 & 47.6 \\
\textbf{MAFormer-S} & 28 & 170 &  \textbf{47.9} & 52  & 929  & \textbf{48.3} & \textbf{48.6}\\
\midrule
Res101~\cite{he2016deep} & 48 & 260 & 38.8  & 86 & 1029 & - & 44.9\\
TwinsP-B ~\cite{2021Twins}  & 48 & 220 & 44.9 & 74  & 977 & 47.1 & 48.4\\
Twins-B ~\cite{2021Twins} & 60 & 261 & 45.3 & 89 & 1020 & 47.7 & 48.9\\
Swin-S ~\cite{liu2021swin} & 53 & 274 & 45.2 & 81 & 1038 & 47.6 & 49.5\\
Focal-S ~\cite{2021Focal} & - & - & - & 85 & 1130 & 48.0 & 50.0 \\
Shuffle-S ~\cite{huang2021shuffle} & - & - & - & 81 & 1044 & 48.4 & 49.6 \\
Swin-B ~\cite{liu2021swin} & 91 & 442 & 46.0  & 121 & 1188 & 48.1 & 49.1 \\
\textbf{MAFormer-B} & 55 & 274 & \textbf{49.8} & 82 & 1031 & \textbf{51.1} & \textbf{51.6} \\
\bottomrule
\hline

\end{tabular}}
\end{center}
\end{table}

\section{Conclusion}
In this paper, we introduce a general vision transformer backbone MAFormer, which integrates local and global features in tokens. MAFormer can improve the information interaction between local windows, where both local and global features are deployed with a linear operation to ensure the consistency of features distribution. With an outstanding performance on image classification and dense downstream tasks, MAFormer has shown its promising potential in vision tasks. In the future, MAFormer can be utilized as a general backbone in the self-supervised pre-training tasks.

\bibliographystyle{ieee}
\bibliography{paper}

\end{document}